# A Large Language Model for Electronic Health Records

**Authors:** Xi Yang[1,2], Aokun Chen[1,2], Nima PourNejatian[3], Hoo Chang Shin[3], Kaleb E Smith[3], Christopher Parisien[3], Colin Compas[3], Cheryl Martin[3], Anthony B Costa[3], Mona G Flores[3], Ying Zhang[4], Tanja Magoc[5], Christopher A Harle[1,5], Gloria Lipori[5,6], Duane A Mitchell[6], William R Hogan[1], Elizabeth A Shenkman[1], Jiang Bian[1,2], Yonghui Wu[1,2] *

**Affiliations:**

[1]Department of Health Outcomes and Biomedical Informatics, College of Medicine, University of Florida, Gainesville, Florida, USA.

[2]Cancer Informatics and eHealth core, University of Florida Health Cancer Center, Gainesville, Florida, USA.

[3]NVIDIA, Santa Clara, California, USA.

[4]Research Computing, University of Florida, Gainesville, Florida, USA.

[5]Integrated Data Repository Research Services, University of Florida, Gainesville, Florida, USA.

[6]Lillian S. Wells Department of Neurosurgery, UF Clinical and Translational Science Institute, University of Florida.

*Corresponding author

Yonghui Wu

Clinical and Translational Research Building

2004 Mowry Road, PO Box 100177, Gainesville, FL, USA, 32610

Phone: 352-294-8436

Email: yonghui.wu@ufl.edu

**Abstract**: There is an increasing interest in developing artificial intelligence (AI) systems to process and interpret electronic health records (EHRs). Natural language processing (NLP) powered by pretrained language models is the key technology for medical AI systems utilizing clinical narratives. However, there are few clinical language models, the largest of which trained in the clinical domain is comparatively small at 110 million parameters (compared with billions of parameters in the general domain). It is not clear how large clinical language models with billions of parameters can help medical AI systems utilize unstructured EHRs. In this study, we develop from scratch a large clinical language model – GatorTron – using >90 billion words of text (including >82 billion words of de-identified clinical text) and systematically evaluate it on 5 clinical NLP tasks including clinical concept extraction, medical relation extraction, semantic textual similarity, natural language inference (NLI), and medical question answering (MQA). We examine how (1) scaling up the number of parameters and (2) scaling up the size of the training data could benefit these NLP tasks. GatorTron models scale up the clinical language model from 110 million to 8.9 billion parameters and improve 5 clinical NLP tasks (e.g., 9.6% and 9.5% improvement in accuracy for NLI and MQA), which can be applied to medical AI systems to improve healthcare delivery. The GatorTron models are publicly available at: https://catalog.ngc.nvidia.com/orgs/nvidia/teams/clara/models/gatortron_og.

**Introduction**

There is an increasing interest in developing artificial intelligence (AI) systems to improve healthcare delivery and health outcomes using electronic health records (EHRs). A critical step is to extract and capture patients' characteristics from longitudinal EHRs. The more information we have about the patients, the better the medical AI systems that we can develop. In recent decades, hospitals and medical practices in the United States (US) have rapidly adopted EHR systems[1,2], resulting in massive stores of electronic patient data, including structured (e.g., disease codes, medication codes) and unstructured (i.e., clinical narratives such as progress notes). Even though using discrete data fields in clinical documentation has many potential advantages and structured data entry fields are increasingly added into the EHR systems, having clinicians use them remains a barrier, due to the added documentation burden[3]. Physicians and other healthcare providers widely use clinical narratives as a more convenient way to document patient information ranging from family medical histories to social determinants of health.[4] There is an increasing number of medical AI systems exploring the rich, more fine-grained patient information captured in clinical narratives to improve diagnostic and prognostic models.[5,6] Nevertheless, free-text narratives cannot be easily used in computational models that usually require structured data. Researchers have increasingly turned to natural language processing (NLP) as the key technology to enable medical AI systems to understand clinical language used in healthcare[7].

Today, most NLP solutions are based on deep learning models[8] implemented using neural network architectures – a fast-developing sub-domain of machine learning. Convolutional neural networks[9] (CNN) and recurrent neural networks[10] (RNN) have been applied to NLP in the early stage of deep learning. More recently, the transformer architectures[11] (e.g., Bidirectional Encoder Representations from Transformers [BERT]) implemented with a self-attention mechanism[12] have become state-of-the-art, achieving the best performance on many NLP benchmarks. [13–16] In the general domain, the transformer-based NLP models have achieved state-of-the-art performance for name entity recognition[17–19], relation extraction[20–24], sentence similarity[25–27], natural language inference[27–30], and question answering[27,28,31,32]. Typically, transformers are trained in two stages: language model pretraining (i.e., learning using a self-supervised training objective on a large corpus of unlabeled text) and fine-tuning (i.e., applying the learned language models solving specific tasks with labeled training data). One pretrained language model can be applied to solve many NLP tasks through fine-tuning, which is known as transfer learning – a strategy to learn knowledge from one task and apply it in another task[33]. Human language has a very large sample space – the possible combinations of words, sentences, and their meaning and syntax are innumerable. Recent studies show that large transformer models trained using massive text data are remarkably better than previous NLP models in terms of emergence and homogenization.[33]

The promise of transformer models has led to further interest in exploring large-size (e.g., >billions of parameters) transformer models. The Generative Pretrained Transformer 3 (GPT-3) model[34], which has 175 billion parameters and was trained using >400 billion words of text demonstrated superior performance. In the biomedical domain, researchers developed BioBERT[11] (with 110 million parameters) and PubMedBERT[35] (110 million parameters) transformer models using biomedical literature from PubMed. NVIDIA developed BioMegatron models in the biomedical domain with different sizes from 345 million to 1.2 billion parameters[36] using a more expansive set of PubMed-derived free text. However, few studies have explored scaling transformer models in the clinical domain due to the sensitive nature of clinical narratives that contain Protected Health Information (PHI) and the significant computing power required to increase the size of these models. To date, the largest transformer model using clinical narratives is ClinicalBERT[37]. ClinicalBERT has 110 million parameters and was trained using 0.5 billion words from the publicly available Medical Information Mart for Intensive Care III[38] (MIMIC-III) dataset. By developing not only larger models, but models that use clinical narratives, NLP may perform better to improve healthcare delivery and patient outcomes.

In this study, we develop a large clinical language model, GatorTron, using >90 billion words of text from the de-identified clinical notes of University of Florida (UF) Health, PubMed articles, and Wikipedia. We train GatorTron from scratch and empirically evaluate how scaling up the number of parameters benefit the performance of downstream NLP tasks. More specifically, we examine GatorTron models with varying number of parameters including (1) a base model with 345 million parameters, (2) a medium model with 3.9 billion parameters, and (3) a large model with 8.9 billion parameters. We also examine how scaling up data size benefit downstream tasks by comparing the GatorTron-base model trained from the full corpus with another GatorTron-base model trained using a random sample of 1/4 of the corpus. We compare GatorTron with existing transformer models trained using biomedical literature and clinical narratives using 5 clinical NLP tasks including clinical concept extraction (or named entity recognition [NER]), medical relation extraction (MRE), semantic textual similarity (STS), natural language inference (NLI), and medical question answering (MQA). GatorTron models outperform previous transformer models from the biomedical and clinical domain on 5 clinical NLP tasks. This study scales up transformer models in the clinical domain from 110 million to 8.9 billion parameters and demonstrates the benefit of large transformer models.

**Results**

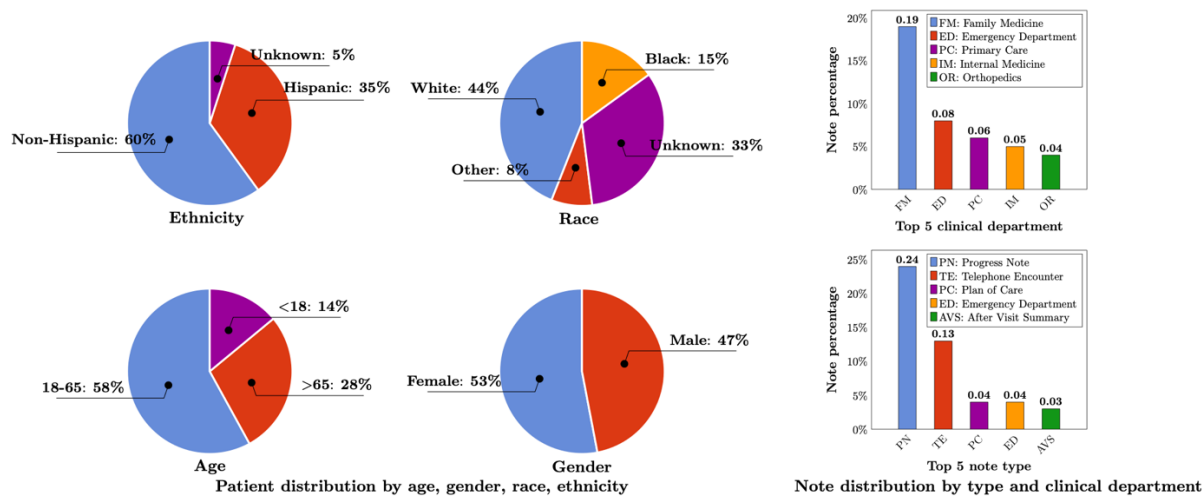

**Fig. 1** Patient distribution by age, gender, race, ethnicity; clinical notes distribution by note type and clinical department. Ages were calculated as of September 2022.

A total number of 290,482,002 clinical notes from 2,476,628 patients were extracted from the UF Health Integrated Data Repository (IDR), the enterprise data warehouse of the UF Health system. These notes were created from 2011-2021 from over 126 clinical departments and approximately 50 million encounters covering healthcare settings including but not limited to inpatient, outpatient, and emergency department visits. After preprocessing and de-identification, the corpus included >82 billion medical words. **Fig. 1** summarizes the distribution of patient by age, gender, race, and ethnicity as well as the distribution of notes by clinical department (top 5) and note type (top 5). The detailed number of patients by each category, a full list of clinical departments and the corresponding proportion of notes, and a full list of note types were provided in **Supplementary Table 1**, **Supplementary Table 2**, and **Supplementary Table 3**.

Training GatorTron-large model required approximately 6 days on 992 A100 80G GPUs from 124 NVIDIA DGX notes using the NVIDIA SuperPOD reference cluster architecture. **Fig. 2** shows the training validation loss for all three sizes of GatorTron models. The GatorTron-base model converged in 10 epochs, whereas the medium and large models converged in 7 epochs, which is consistent with prior observations on the faster per sample convergence of larger transformer models.

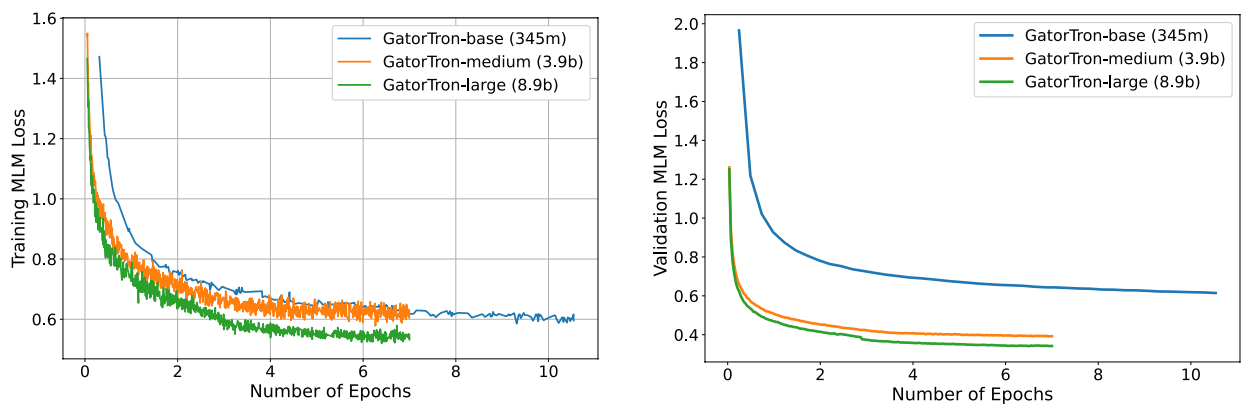

**Fig. 2** Training loss and validation loss for GatorTron base (345 million), medium (3.9 billion), and large (8.9 billion) models.

**Table 1.** Comparison of GatorTron with existing biomedical and clinical transformer models for clinical concept extraction and medical relation extraction.

|  | Clinical concept extraction | | | | | | | | | Medical relation extraction | | |
|---|---|---|---|---|---|---|---|---|---|---|---|---|
|  | 2010 i2b2[39] | | | 2012 i2b2[40] | | | 2018 n2c2[41] | | | 2018 n2c2[41] | | |
| **Transformer** | Precision | Recall | F1 score | Precision | Recall | F1 score | Precision | Recall | F1 score | Precision | Recall | F1 score |
| BioBERT | 0.8693 | 0.8653 | 0.8673 | 0.7478 | 0.8037 | 0.7747 | 0.8634 | 0.8921 | 0.8775 | 0.9663 | 0.9451 | 0.9555 |
| ClinicalBERT | NA | NA | 0.8780 | NA | NA | 0.7890 | 0.8592 | 0.8832 | 0.8710 | 0.9678 | 0.9414 | 0.9544 |
| BioMegatron | 0.8614 | 0.8761 | 0.8687 | 0.7591 | 0.8031 | 0.7805 | 0.8707 | 0.8915 | 0.8810 | 0.9711 | 0.9434 | 0.9571 |
| GatorTron-base (1/4 data) | 0.8682 | 0.9046 | 0.8860 | 0.7514 | 0.8013 | 0.7755 | 0.8772 | 0.8992 | 0.8881 | 0.9724 | 0.9457 | 0.9589 |
| GatorTron-base | 0.8748 | 0.9043 | 0.8893 | 0.7644 | 0.8221 | 0.7922 | 0.8759 | 0.9038 | 0.8896 | 0.9719 | 0.9482 | 0.9599 |
| GatorTron-medium | 0.8869 | 0.9122 | 0.8994 | 0.7812 | 0.8245 | 0.8022 | 0.8954 | 0.9035 | 0.8994 | 0.9721 | 0.9503 | 0.9611 |
| GatorTron-large | 0.8880 | 0.9116 | **0.8996** | 0.7862 | 0.8333 | **0.8091** | 0.8979 | 0.9021 | **0.9000** | 0.9776 | 0.9482 | **0.9627** |

Clinical concepts in 2010 i2b2 and 2012 i2b2 challenges: problems, treatments, lab tests; clinical concepts in 2018 n2c2 challenge: drugs, adverse events, and drug-related attributes (e.g., dose). Medical relation in 2018 n2c2 challenge: drug induced adverse events. Best F1 scores are bolded. NA: scores not reported.

**Table 2.** Comparison of GatorTron with existing biomedical and clinical transformer models for semantic textual similarity, natural language inference, and question answering.

| | Semantic textual similarity | Natural language inference | Question answering | | | |
|---|---|---|---|---|---|---|
| | 2019 n2c2[42] | MedNLI[43] | emrQA Medication[44] | | emrQA Relation[44] | |
| **Transformer** | Pearson correlation | Accuracy | F1 score | Exact Match | F1 score | Exact Match |
| BioBERT | 0.8744 | 0.8050 | 0.6997 | 0.2475 | 0.9262 | 0.8361 |
| ClinicalBERT | 0.8787 | 0.8270 | 0.6905 | 0.2406 | 0.9306 | 0.8533 |
| BioMegatron | 0.8806 | 0.8390 | 0.7231 | 0.2882 | 0.9405 | 0.879 |
| GatorTron-base (1/4 data) | 0.8675 | 0.8643 | 0.7281 | 0.2952 | 0.9390 | 0.8579 |
| GatorTron-base | 0.8810 | 0.8670 | 0.7181 | 0.2978 | 0.9543 | 0.9029 |
| GatorTron-medium | **0.8903** | 0.8720 | 0.7354 | 0.3018 | 0.9677 | 0.9243 |
| GatorTron-large | 0.8896 | **0.9020** | **0.7408** | **0.3155** | **0.9719** | **0.9310** |

The best evaluation scores are bolded.

**Table 1** and **Table 2** compare GatorTron models with two existing biomedical transformer models (BioBERT and BioMegatron) and one clinical transformer model (Clinical BERT) on 5 clinical NLP tasks.

**Scale up the size of training data and the number of parameters**. Compared with GatorTron-base trained using a random sample of 1/4 of the corpus, the GatorTron-base model trained using the full corpus achieved improved performance for 4 tasks except for a sub-task in MQA (on F1 score of medication-related questions). By scaling up the number of parameters from 345 million to 8.9 billion, GatorTron-large demonstrated remarkable improvements for all 5 tasks, suggesting that GatorTron models scale for canonical clinical downstream tasks and that we are not yet at the limit.

**Recognize clinical concepts and medical relations**. Clinical concept extraction is to identify the concepts with important clinical meanings and classify their semantic categories (e.g., diseases, medications). As shown in Table 1, all three GatorTron models outperformed existing biomedical and clinical transformer models in recognizing various types of clinical concepts on the three benchmark datasets (i.e., 2010 i2b2[39] and 2012 i2b2[40]: problem, treatments, lab tests; 2018 n2c2[41]: drug, adverse events, and drug-related attributes). The GatorTron-large model outperformed the other two smaller GatorTron models and achieved the best F1-scores of 0.8996, 0.8091, and 0.9000, respectively. For medical relation extraction – a task to identify medical relations between two clinical concepts – the GatorTron-large model also achieved the best F1-score of 0.9627 for identifying drug-cause-adverse event relations outperforming existing biomedical and clinical transformers and the other two smaller GatorTron models. We consistently observed performance improvement when scaling up the size of the GatorTron model.

**Assess semantic textual similarity**. The task of measuring semantic similarity is to determine the extent to which two sentences are similar in terms of semantic meaning. As shown in Table 2, all GatorTron models outperformed existing biomedical and clinical transformer models. Among the three GatorTron models, the GatorTron-medium model achieved the best Pearson correlation score of 0.8903, outperforming both GatorTron-base and GatorTron-large. Although we did not observe consistent improvement by scaling up the size of the GatorTron model, the

GatorTron-large model outperformed GatorTron-base and its performance is very close to the GatorTron-medium model (0.8896 vs. 0.8903).

**Natural language inference.** The task of NLI is to determine whether a conclusion can be inferred from a given sentence – a sentence-level NLP task. As shown in Table 2, all GatorTron models outperformed existing biomedical and clinical transformers, and the GatorTron-large model achieved the best accuracy of 0.9020, outperforming the BioBERT and ClinicalBERT by 9.6% and 7.5%, respectively. We observed a monotonic performance improvement by scaling up the size of the GatorTron model.

**Medical question answering**. MQA is a complex clinical NLP task that requires understand information from the entire document. As shown in Table 2, all GatorTron models outperformed existing biomedical and clinical transformer models in answering medication and relation-related questions (e.g., "What lab results does patient have that are pertinent to diabetes diagnosis?"). For medication-related questions, the GatorTron-large model achieved the best exact match score of 0.3155, outperforming the BioBERT and ClinicalBERT by 6.8% and 7.5%, respectively. For relation-related questions, GatorTron-large also achieved the best exact match score of 0.9301, outperforming BioBERT and ClinicalBERT by 9.5% and 7.77%, respectively. We also observed a monotonic performance improvement by scaling up the size of the GatorTron model.

## Discussion

In this study, we developed a large clinical transformer model, GatorTron, using a corpus of >90 billion words from UF Health (>82 billion), Pubmed (6 billion), Wikipedia (2.5 billion), and MIMIC III (0.5 billion). We trained GatorTron with different number of parameters including

345 million, 3.9 billion, and 8.9 billion and evaluated its performance on 5 clinical NLP tasks at different linguistic levels (phrase level, sentence level, and document level) using 6 publicly available benchmark datasets. The experimental results show that GatorTron models outperformed existing biomedical and clinical transformers for all 5 clinical NLP tasks evaluated using 6 different benchmark datasets. We observed monotonic improvements by scaling up the model size of GatorTron for 4 of the 5 tasks, excluding the semantic textual similarity task. Our GatorTron model also outperformed the BioMegatron[36], a transformer model with a similar model size developed in our previous study using >8.5 billion words from PubMed and Wikipedia (a small proportion of the >90 billion words of corpus for developing GatorTron). This study scaled up the clinical transformer models from 345 million (ClinicalBERT) to 8.9 billion parameters in the clinical domain and demonstrated remarkable performance improvements. To the best of our knowledge, GatorTron-large is the largest transformer model in the clinical domain. Among the 5 tasks, GatorTron achieved remarkable improvements for complex NLP tasks such as natural language inference and medical question answering, but moderate improvements for easier tasks such as clinical concept extraction and medical relation extraction, indicating that large transformer models are more helpful to complex NLP tasks. These results are consistent with observations in the literature on the saturation of simpler benchmarks with large BERT architectures.[18,32]

GatorTron was pretrained using self-supervised masked language modeling (MLM) objective. We monitored training loss and calculated validation loss using a subset set of the clinical text (5%) to determine the appropriate stopping time. From the plots of training and validation losses in Fig. 2, we observed that larger GatorTron models converged faster than the smaller model.

GatorTron models perform better in extracting and interpreting patient information documented in clinical narratives, which can be integrated into medical AI systems to improve healthcare delivery and patient outcomes. The rich, fine-grained patient information captured in clinical narratives is a critical resource powering medical AI systems. With better performance in information extraction (e.g., clinical concept extraction and medical relation extraction), GatorTron models can provide more accurate patient information to identify research-standard patient cohorts using computable phenotypes, support physicians making data-informed decisions by clinical decision support systems, and identify adverse events associated with drug exposures for pharmacovigilance. The observed improvements in semantic textual similarity, natural language inference, and medical question answering can be applied for deduplication of clinical text, mining medial knowledge, and developing next-generation medical AI systems that can interact with patients using human language.

We conducted error analysis and compared GatorTron with ClinicalBERT to probe the observed performance improvements. We found that the larger, domain-specific pre-trained models (e.g., GatorTron) are better at modeling longer phrases and determining semantic categories. For example, GatorTron successfully identified "*a mildly dilated ascending aorta*", where ClinicalBERT identified only "mildly dilated" as a problem; GatorTron successfully categorized "kidney protective effects" as a "TREATMENT", which was mis-classified as "PROBLEM" by ClinicalBERT. For complex NLP tasks such as NLI and MQA, even large language models such as GatorTron still have difficulty in identifying the key pieces of information from longer paragraphs. Our future work will improve GatorTron in handling long pieces of text for complex NLP tasks.

This study demonstrates the advantages of large pretrained transformer models in the medical domain. GatorTron models can be applied to many other NLP tasks through fine-tuning. We believe that GatorTron will improve the use of clinical narratives in developing various medical AI systems for better healthcare delivery and health outcomes.

**Methods**

**Data Source**

The primary data source for this study is the clinical narratives from UF Health IDR, a research data warehouse of UF Health. This study was approved by the UF Institutional Review Board (IRB202100049). We collected clinical notes from 2011-2021 from over 126 departments, approximately 2 million patients and 50 million encounters from inpatient, outpatient, and emergency settings. Then, we merged the UF Health clinical corpus with three additional corpora, including the MIMIC-III corpus[38] in the clinical domain with 0.5 billion words, a PubMed (combining PubMed abstracts and full-text commercial-collection) collection[36] in the biomedical domain with 6 billion words, and a Wikipedia articles dump[36] in the general domain with 2.5 billion words, to generate a corpus with >90 billion words.

**Preprocessing and de-identification of text**

We performed minimal preprocessing including (1) removing empty and duplicated clinical notes, unifying all text into UTF-8 encoding, and removing illegal UTF-8 strings; (2) normalizing special characters (e.g., convert '&' to '&;' '\xa0' to 'space'); (3) tokenization and sentence boundary detection. For clinical text from UF Health, we further applied a de-identification system[45] to remove protected health information (PHI) from clinical text. (Approved under IRB202100049) We adopted the safe-harbor method to identify 18 PHI categories defined in the Health Insurance Portability and Accountability Act (HIPAA) and replaced them with dummy strings (e.g., replace people's names into [**NAME**]).

**Study design**

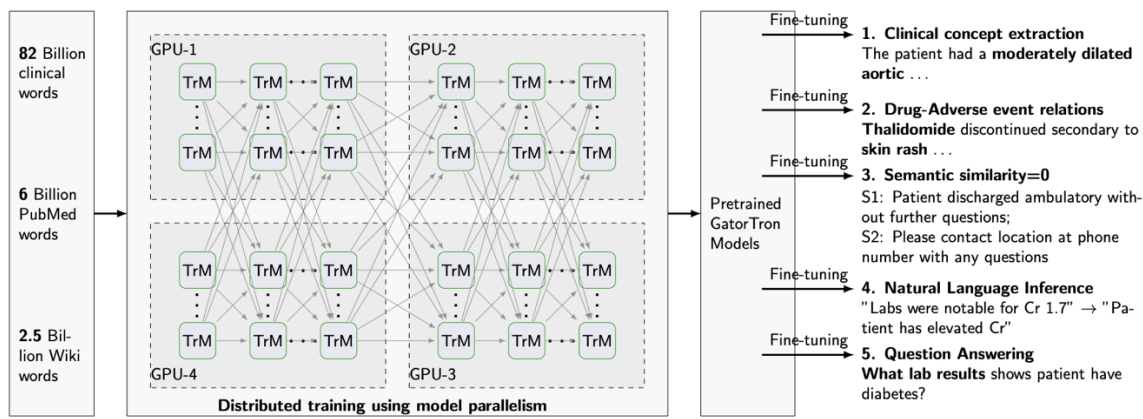

**Fig. 3** An overview of pretraining and fine-tuning of GatorTron models. We loaded the base model and the medium model into one GPU for distributed training. We sliced the GatorTron-large model into 4 pieces and loaded model pieces to 4 GPUs for distributed training (i.e., model parallelism). TrM: transformer unit.

**Fig. 3** shows an overview of the study design. We seek to train a large clinical transformer model, GatorTron, using >90 billion words and examine how and whether scaling up model size improves performance on 5 clinical NLP tasks. We first pretrained GatorTron using the >90 billion words by optimizing a masked language model (MLM) and then applied GatorTron to 5 different clinical NLP tasks using a supervised fine-tuning. We adopted the BERT architecture (**Fig. 4**) implemented in Megatron-LM and explored three different settings including a base model of 345 million parameters (i.e., GatorTron-base), a medium model of 3.9 billion parameters (i.e., GatorTron-medium), and a large model of 8.9 billion parameters (i.e., GatorTron-large). Then we compared the three GatorTron models to an existing transformer model from the clinical domain, ClinicalBERT (trained with 110 million parameters) and two transformer models from the biomedical domain, including, BioBERT (345 million parameters) and BioMegatron (1.2 billion parameters). We compared the models on 5 clinical NLP tasks, including clinical concept extraction, relation extraction, semantic textual similarity, natural language inference, and medical question answering. We used 6 public benchmark datasets in the clinical domain.

**Training environment**

We used a total number of 992 NVIDIA DGX A100 GPUs from 124 superPOD nodes at UF's HiPerGator-AI cluster to train GatorTron models by leveraging both data-level and model-level parallelisms implemented by the Megatron-LM package[46]. We monitored the training progress by training loss and validation loss and stopped the training when there was no further improvement (i.e., the loss plot became flat).

**GatorTron Model Configuration**

We developed GatorTron models with three configurations and determined the number of layers, hidden sizes, and number of attention heads according to the guidelines for optimal depth-to-width parameter allocation proposed by Levin et al[47] as well as our previous experience in developing BioMegatron. **Table 3** provides detailed information for the three settings. The GatorTron-base model has 24 layers of transformer blocks, which is similar to the architecture of BERT large model. For each layer, we set the number of hidden units as 1024 and attention heads as 16. The GatorTron-medium model scaled up to 3.9 billion parameters (~10 times of the base setting) and the GatorTron-large model scaled up to 8.9 billion parameters, which is similar to BioMegatron[46] ( with 8.3 billion parameters).

**Table 3.** Technical details of GatorTron models.

| Model | # Layers | # Hidden Size | # Attention Heads | # Parameters |
| --- | --- | --- | --- | --- |
| GatorTron-base | 24 | 1024 | 16 | 345 million |
| GatorTron-medium | 48 | 2560 | 40 | 3.9 billion |
| GatorTron-large | 56 | 3584 | 56 | 8.9 billion |

**Train GatorTron models from scratch**

We pretrained a vocabulary from scratch using >90 billion words of corpus following the byte-pair-encoding algorithm.[48] We inherited the BERT-style architecture and trained GatorTron models from scratch using two self-supervised tasks, including masked language modeling (MLM) and sentence-order prediction (SOP). We followed the similar strategy in the BERT model[49] to randomly mask 15% of the input tokens with a special token (i.e., [MASK]) in the MLM. The SOP was formulated as a task to predict the order of two consecutive segments of text.[28] The input for SOP consists of two consecutive sentences from the training corpus in random orders and the training objective is to determine whether the two input sentences are in the correct order. The GatorTron large model with 8.9 billion parameters is too large to fit one GPU, therefore, we sliced it into 4 pieces for distributed training using model parallelism. We pretrained the GatorTron base and medium model without model slicing. The default loss function defined in BERT model[49] was used. **Fig. 4** shows the distributed training of GatorTron large model using model parallelism. (See https://github.com/NVIDIA/Megatron-LM for more details)

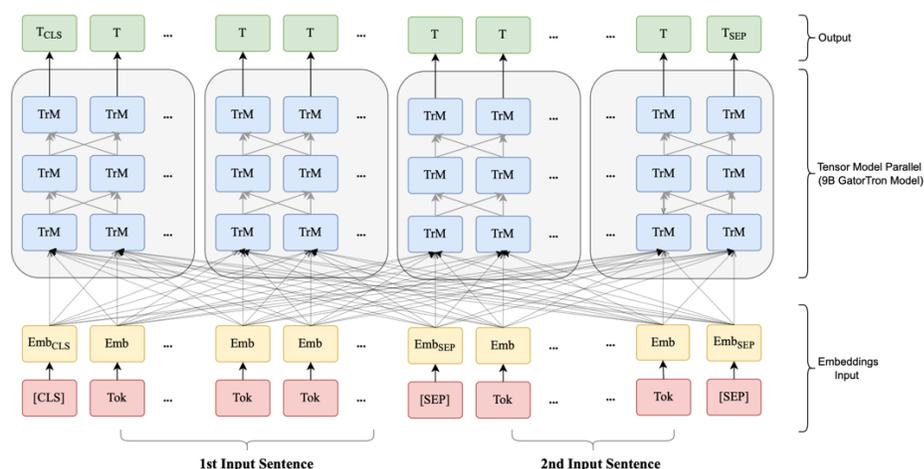

**Fig. 4** Pretraining GatorTron large model with 9 billion parameters using model parallelism. Emb: embedding; Tok: Token from input sentence; Trm: Transformer unit. [SEP]: a token defined in BERT to indicate sentence boundaries. [CLS]: a token defined in BERT for sentence-level representation.

**Existing transformer models for comparison**

**BioBERT.**[11] The BioBERT model was developed by further training the original BERT-large model (345 million parameters, 24 layers, 1024 hidden units, and 16 attention heads) using biomedical literature from PubMed Abstracts (4.5 billion words) and PMC Full-text articles (13.5 billion words). In this study, we used version 1.1.

**ClinicalBERT**.[37] The ClinicalBERT model was developed by further training the BioBERT (base version; 110 million parameters with 12 layers, 768 hidden units, and 12 attention heads) using clinical text from the MIMIC-III[38] corpus.

**BioMegatron**.[36] The BioMegatron models adopted the BERT architecture with a different number of parameters from 345 million to 1.2 billion. Different from BioBERT and ClinicalBERT, the BioMegatron was trained from scratch without leveraging the original BERT model.

**Fine-tune GatorTron for 5 clinical NLP tasks, evaluation matrices, and benchmark datasets**

We fine-tuned pretrained GatorTron models for 5 different clinical NLP tasks using experts' annotations from 6 public benchmark datasets. Specifically, we first generated distributed representation from the inputs of a specific task, then added additional output layers (classification or regression) to generate target outputs. We used cross-entropy (CE) loss for classification tasks and mean square error loss for regression tasks. For a classification task with *N* categories, let $C_i$ be the score generated by a transformer model for category *i*, the probability $P_i$ of a given sample be classified to category *i* was calculated as:

$$P_i = \frac{e^{C_i}}{\sum_{j=1}^{N} e^{C_j}} \tag{1}$$

Let $t_i$ be the ground truth category, the cross-entropy loss $L_{CE}$ is defined as:

$$L_{CE} = -\sum_{i=1}^{N} t_i \log(P_i) \tag{2}$$

**Fine-tune GatorTron for Clinical Concept Extraction.** This is a task to recognize phrases with important clinical meanings (e.g., medications, treatments, adverse drug events). The task is to determine the boundaries of a concept and classify it into predefined semantic categories. Early systems for clinical concept extract are often rule-based, yet, most recent systems are based on machine learning models such as conditional random fields (CRFs)[50,51], convolutional neural networks (CNN)[9,52], and recurrent neural networks (RNN) implemented with long short-term memory strategy (LSTM)[10,53]. Current state-of-the-art models are based on transformers such as the ClinicalBERT. We approached clinical concept extraction as a sequence labeling problem and adopted 'BIO' labeling schema, where 'B-' and 'I-' are prefixes indicating words at the beginning and inside of a concept, and 'O' stands for words located outside of any concepts of interest. Using this definition, we approached the task as a classification problem – for each word in a sentence, predict a label in ['B', 'I', 'O']. When there are multiple categories of concepts, a suffix was attached to 'BIO' for discrimination (e.g., 'B-drug', 'I-drug'). Based on the representation generated by pretrained GatorTron models, we added a classification layer (a linear layer with softmax activation) to calculate a probability score for each 'BIO' category. The cross-entropy loss was used for fine-tuning. We trained a unified classifier to extract all concepts for datasets without overlapped concepts. For datasets with overlapped concepts, we trained individual models to recognize each category of concept separately following our previous strategy. [54] We used three benchmark datasets developed by the 2010 i2b2 challenge[39], 2012 i2b2 challenge[40], and 2018 n2c2 challenge[41] to evaluate GatorTron models focusing on identifying important medical concepts (e.g., medications, adverse drug events, treatments) from clinical text. We used precision, recall, and F1-score for evaluation.

**Fine-tune GatorTron for Medical Relation Extraction.** MRE is to establish medical-related relations (e.g., induce relation) among clinical concepts (e.g., drugs, adverse events). MRE is usually approached as a classification problem – identify pairs of concepts with valid relations and classify the relation type. Various machine learning-based classifiers such as support vector machines (SVMs), random forests (RF), and gradient boosting trees (GBT)[41] have been applied. With the emergence of deep learning models, researchers have explored the long-short term memory (LSTM) architecture for RE in both general and clinical domains[55,56]. Most recently, several studies adopted the BERT architecture and demonstrated superior performance for MRE on various datasets[57–62]. We approached MRE as a classification task. First, candidate concept pairs were generated using heuristic rules developed in our previous study[41]. Then, we identified two sentences where the two concepts in a pair were located. We introduced two sets of entity markers (i.e., [S1], [E1] and [S2], [E2]) to indicate the two concepts. If the two concepts were in the same sentence, the two input sentences will be the same but labeled with different markers (e.g., [S1] and [E1] were used in the first sentence; [S2] and [E2] were used in the second sentence). To determine the relation type, we concatenated the representations of the model special [CLS] token and all four entity markers and added a classification layer (a linear layer with softmax activation) for classification. Similarly, the cross-entropy loss was used to fine-tune GatorTron. We used the dataset developed by the 2018 n2c2 challenge[41] with a focus on relations between medications and adverse drug events. The precision, recall, and F1-score were used for evaluation.

**Fine-tune GatorTron for Semantic Textual Similarity.** The STS task is to quantitatively assess the semantic similarity between two text snippets (e.g., sentences), which is usually approached as a regression task where a real-value score was used to quantify the similarity between two text snippets. In the general domain, the STS benchmark (STS-B) dataset curated by the Semantic Evaluation (SemEval) challenges between 2012 and 2017[63] is widely used for evaluating STS systems[13]. Various machine learning methods have been examined[64–66] but transformer-based systems such as RoBERTa[25], T5[27], and ALBERT[28] are leading the state-of-the-art models for STS. In the clinical domain, the MedSTS dataset[67] that consists of over 1000 annotated sentence pairs from clinical notes at Mayo Clinic was widely used as the benchmark. MedSTS was used as the gold standard in two clinical NLP open challenges including the 2018 BioCreative/Open Health NLP (OHNLP) challenge[68] and 2019 n2c2/OHNLP ClinicalSTS shared task[42]. Similar to the general domain, pretrained transformer-based models using clinical text and biomedical literature, including ClinicalBERT and BioBERT[69], achieved state-of-the-art performance. In this study, we formulated STS as a regression problem. We applied pretrained GatorTron models to learn the sentence-level representations of the two pieces of text and adopted a linear regression layer to calculate the similarity score. Different from classification models, we used MSE as the loss function. We used the dataset developed by the 2019 n2c2/OHNLP[42] challenge on clinical semantic textural similarity[42]. The Pearson correlation score was used for evaluation.

**Fine-tune GatorTron for Natural Language Inference**. NLI is also known as recognizing textual entailment (RTE) - a directional relation between text fragments (e.g., sentences)[70]. The goal of NLI is to determine if a given hypothesis can be inferred from a given premise. In the general domain, two benchmark datasets - the MultiNLI[71] and the Stanford NLI[72] are widely used. On both datasets, pretrained transformer models achieved state-of-the-art performances[27,29]. There are limited resources for NLI in the clinical domain. Until recently, the MedNLI – a dataset annotated by doctors based on the medical history of patients[43] was developed as a benchmark dataset in the clinical domain. A previous study[37] showed that a pretrained clinical BERT model achieved the state-of-the-art performance and outperformed the baseline (InferSent[73]) by ~9% accuracy. In this study, we approached NLI as a classification problem. We concatenated the hypothesis and premise as the input separated using a special token [SEP] and applied pretrained GatorTron models to generate distributed representations, which were fed into a classification layer (a linear layer with softmax activation) to calculate a probability for each of the three categories of entailment, contradiction, and neutral. The cross-entropy loss was used for fine-tuning. We evaluated the GatorTron models on NLI using the MedNLI dataset[43] and used accuracy for comparison.

**Fine-Tune GatorTron for Medical Question Answering**. The MQA task is to build NLP systems that automatically answer medical questions in a natural language, which is the most complex challenge among the 5 tasks. Unlike other tasks focusing on phrases and sentences,

MQA is a document-level task that requires information from the whole document to generate answers according to questions. In the general domain, the Stanford Question Answering Datasets (SQuAD 1.1 and 2.0)[74,75] have been widely used as benchmarks. Transformer-based models are state-of-the-art for both SQuAD1.1[18] and SQuAD2.0[31]. There are several MQA datasets developed in the past few years such as the MESHQA[76], MedQuAD[77], and emrQA[44]. In this study, we approached MQA using a machine reading comprehension (MRC) technique where the goal is to extract the most relevant responses (i.e., short text snippets or entities) from the given context according to questions. We applied a span classification algorithm to identify the start and end offsets of the answer from the context. More specifically, we packed the question and the context into a single sequence as input for GatorTron and applied two linear layers to predict the start and end position of the answer, respectively. As GatorTron models were developed using a maximum token length of 512, we limited the maximum length of questions to 64 tokens and the rest of the 446 tokens (including special tokens such as [CLS] and [SEP]) were used for the context. We truncated questions with more than 64 tokens. For contexts the had more than 446 tokens, we adopted a sliding window strategy to scan the whole document using a window size of 446 tokens and a stride size of 396 tokens, so that two consecutive windows had the same 50 tokens overlapped. We also limited the answers to a maximum length of 32 tokens. We used the emrQA dataset[44], which is widely used as a benchmark dataset for MQA. We particularly focused on medications and relations-related questions as Yue et al.[78] found that the two subsets are more consistent. We utilized both F1-score and exact match score for evaluation.

**Data availability**

The benchmark datasets that support the findings of this study are available from the official websites of natural language processing challenges with Data Use Agreements. More specifically:

1. i2b2 2010, 2012 datasets and n2c2 2018, 2019 datasets: https://portal.dbmi.hms.harvard.edu/projects/n2c2-nlp/

2. MedNLI dataset: https://physionet.org/content/mednli/1.0.0/

3. emrQA dataset: https://github.com/panushri25/emrQA#download-dataset

4. MIMIC III dataset: https://physionet.org/content/mimiciii/1.4/

5. PubMed dataset: https://www.ncbi.nlm.nih.gov/pmc/tools/openftlist/

6. Wikipedia dataset: https://dumps.wikimedia.org/enwiki/latest/enwiki-latest-pages-articles.xml.bz2

7. UF Health IDR clinical notes are not open to the public due to patient privacy information. The GatorTron models pretrained using >90 billion words of text is publicly available at: https://catalog.ngc.nvidia.com/orgs/nvidia/teams/clara/models/gatortron_og

**Code Availability**

The computer codes to train GatorTron models are available from: https://github.com/NVIDIA/Megatron-LM and https://github.com/NVIDIA/NeMo

The computer codes for preprocessing of text data are available from:

https://github.com/uf-hobi-informatics-lab/NLPreprocessing
https://github.com/uf-hobi-informatics-lab/GatorTron

**Acknowledgments:**

This study was partially supported by a Patient-Centered Outcomes Research Institute® (PCORI®) Award (ME-2018C3-14754), a grant from the National Cancer Institute, 1R01CA246418 R01, grants from the National Institute on Aging, NIA R56AG069880 and R21AG062884, and the Cancer Informatics and eHealth core jointly supported by the UF Health Cancer Center and the UF Clinical and Translational Science Institute. The content is solely the responsibility of the authors and does not necessarily represent the official views of the funding institutions.

We would like to thank the UF Research Computing team, led by Dr. Erik Deumens, for providing computing power through UF HiPerGator-AI cluster.

**Author contributions**

YW, JB, MGF, NP, and XY were responsible for the overall design, development, and evaluation of this study. XY and AC had full access to all the data in the study and takes responsibility for the integrity of the data and the accuracy of the data analysis. YW, XY JB, and WH did the bulk of the writing, EAS, DAM, TM, CAH, ABC, and GL also contributed to writing and editing of this manuscript. All authors reviewed the manuscript critically for scientific content, and all authors gave final approval of the manuscript for publication.

**Competing interests**
The Authors declare no Competing Financial or Non-Financial Interests.

**Legends of Figures**

**Fig. 1** Patient distribution by age, gender, race, ethnicity; clinical notes distribution by note type and clinical department. Ages were calculated as of September 2022.

**Fig. 2** Training loss and validation loss for GatorTron base (345 million), medium (3.9 billion), and large (8.9 billion) models.

**Fig. 3** An overview of pretraining and fine-tuning of GatorTron models. We loaded the base model and the medium model into one GPU for distributed training. We sliced the GatorTron-large model into 4 pieces and loaded model pieces to 4 GPUs for distributed training (i.e., model parallelism). TrM: transformer unit.

**Fig. 4** Pretraining GatorTron large model with 9 billion parameters using model parallelism. Emb: embedding; Tok: Token from input sentence; Trm: Transformer unit. [SEP]: a token defined in BERT to indicate sentence boundaries. [CLS]: a token defined in BERT for sentence-level representation.

**Legends of Tables**

**Table 1.** Comparison of GatorTron with existing biomedical and clinical transformer models for clinical concept extraction and medical relation extraction.

**Table 2.** Comparison of GatorTron with existing biomedical and clinical transformer models for semantic textual similarity, natural language inference, and question answering.

**Table 3.** Technical details of GatorTron models.

**Materials & Correspondence**

Yonghui Wu

Clinical and Translational Research Building

2004 Mowry Road, PO Box 100177, Gainesville, FL, USA, 32610

Phone: 352-294-8436

Email: yonghui.wu@ufl.edu

**Statistical information: N/A**